\documentclass{bmvc2k}
\usepackage{amsmath}
\usepackage{amssymb}
\usepackage{float}
\usepackage{multirow}
\usepackage[nobiblatex]{xurl}

\title{Generate Your Own Scotland: Satellite Image Generation Conditioned on Maps}

\addauthor{Miguel Espinosa}{miguel.espinosa@ed.ac.uk}{1}
\addauthor{Elliot J. Crowley}{elliot.j.crowley@ed.ac.uk}{1}

\addinstitution{
 School of Engineering\\
 University of Edinburgh
}

\runninghead{Espinosa and Crowley}{Generate Your Own Scotland}

\begin{document}

\maketitle

\begin{abstract}

Despite recent advancements in image generation, diffusion models still remain largely underexplored in Earth Observation. In this paper we show that state-of-the-art pretrained diffusion models can be conditioned on cartographic data to generate realistic satellite images. We provide two large datasets of paired OpenStreetMap images and satellite views over the region of Mainland Scotland and the Central Belt. We train a ControlNet model and qualitatively evaluate the results, demonstrating that both image quality and map fidelity are possible. Finally, we provide some insights on the opportunities and challenges of applying these models for remote sensing. Our model weights and code for creating the datasets are publicly available at~\url{https://github.com/miquel-espinosa/map-sat}.

\end{abstract}

\begin{figure*}[h]
\begin{center}
\includegraphics[scale=0.848]{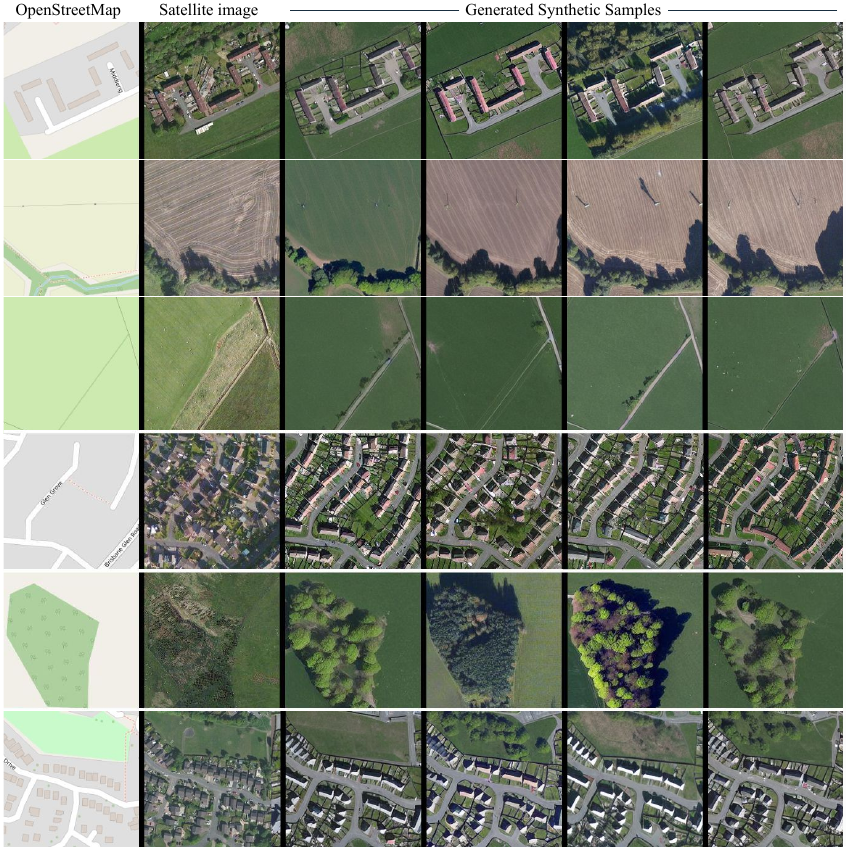}
\end{center}
   \caption{Examples of synthetic sat. images generated with diffusion models conditioned on OSM maps (test set). The real sat. images are provided as reference (2nd column) but they are not used at inference. We cover a wide landscape diversity (urbanised and rural areas).}
\label{fig:examples-clarity}
\end{figure*}

\section{Introduction}
Earth Observation (EO) is a rapidly expanding field that uses computer vision, machine learning, and image processing to gain insights into Earth's surface changes~\cite{Kansakar2016May, Persello2021Dec}. For this purpose, it is crucial to extract meaningful information from diverse and often noisy data sources. Recently, the use of maps in EO has gained attention due to their abstract representation capabilities \cite{Audebert2017May, Wu2020Jul, Vargas_Munoz2020Jun}. However, the use of cartographic data in remote sensing still remains largely unexplored. Maps, such as OpenStreetMap (OSM) \cite{OpenStreetMap}, offer high-quality information about roads, buildings, railways, and more, that can enhance EO analyses when paired with satellite images~\cite{Li2022Jun, Li2021Dec, Li2020Aug}.

Generative models, specifically diffusion models, have shown great potential in different sectors, e.g. for medical imaging~\cite{Kazerouni2023Aug}. As the quality of the generated images improves, it is important for the remote sensing community to adopt these methods, not only for their ability to augment datasets and create realistic synthetic images but also for their implications in distinguishing real from artificially generated or manipulated content \cite{Zhao2021Jul}, thus, promoting its responsible and ethical use.

Our work combines OSM maps and generative models to synthesise realistic satellite views. We make two main contributions. First, we create a large dataset of image pairs, combining map data and satellite imagery. With this dataset, we highlight the importance of using different data sources in EO; specifically, we demonstrate new possibilities when using cartographic data. Second, we show that advanced generative models can be used effectively in the domain of satellite imagery. For this, we train ControlNet \cite{Zhang2023Feb}, a state-of-the-art model, to generate high-quality high-fidelity images. We demonstrate how this generation can be controlled and conditioned with different input data, such as maps. With this study, we hope to spark new research interests in this direction.

\section{Background}

Generative models have significantly improved in recent years. Several works have explored their use for synthetic image generation \cite{Radford2015Nov}, image-to-image translation \cite{Isola2016Nov}, and data augmentation \cite{Antoniou2017Nov}. However, in the EO domain the focus has predominantly been on more traditional models, such as Generative Adversarial Networks (GANs) \cite{Goodfellow2014Jun}. While GANs have shown notable results in multiple EO tasks (super resolution \cite{Demiray2021Feb, Sun2020Jan}, de-speckling \cite{Wang2017Dec}, pan-sharpening \cite{Ma2020Oct}, image generation \cite{Le2023Feb}, haze or cloud removal \cite{Hu2020Dec}), they suffer from training instability and model collapse, which can lead to the generation of low quality images~\cite{Arjovsky2017Jan}.

Recent developments in generative modeling have opened new avenues for research. Particularly, diffusion models \cite{Sohl_Dickstein2015Mar, Ho2020Jun} have emerged as promising alternatives, using stochastic processes to model the data distribution. Previous work has explored the use of diffusion models in the EO domain for diverse downstream applications such as super resolution~\cite{Liu2022Sep}, change detection~\cite{Bandara2022Jun}, and image augmentation~\cite{Adedeji2022Jul}. Recent work such as ControlNet \cite{Zhang2023Feb} allows for better control over the generation process by adding input conditions while still produce high-quality results. The use of such conditioned diffusion models in remote sensing remains unexplored, creating a gap that our work aims to address.

In the multi-modal context, previous work has explored the use of paired datasets combining different types of remote sensing data \cite{Wu2021Nov, Sharma2020Nov, Hong2020Aug, Li2022Aug}. However, the use of cartographic maps as an additional data source remains underexplored.

\section{Datasets}
\label{sec:dataset}

To demonstrate the effectiveness of pretrained diffusion models in remote sensing, we construct a specific dataset for the training procedure. Instructions for the dataset creation and the code used can be found in~\url{https://github.com/miquel-espinosa/map-sat}.

The multi-modal dataset pairs $256\times 256$ OSM image tiles with corresponding $256\times 256$ World Imagery \cite{worldimagery} satellite image tiles. We use a fixed text prompt ``\emph{Convert this OpenStreetMap into its satellite view}'' for the pretrained SD model. The area considered in this study is mainland Scotland. The sampling strategy consists of random sampling over a predefined region.

\begin{figure*}[h]
\begin{center}
\includegraphics[scale=0.5]{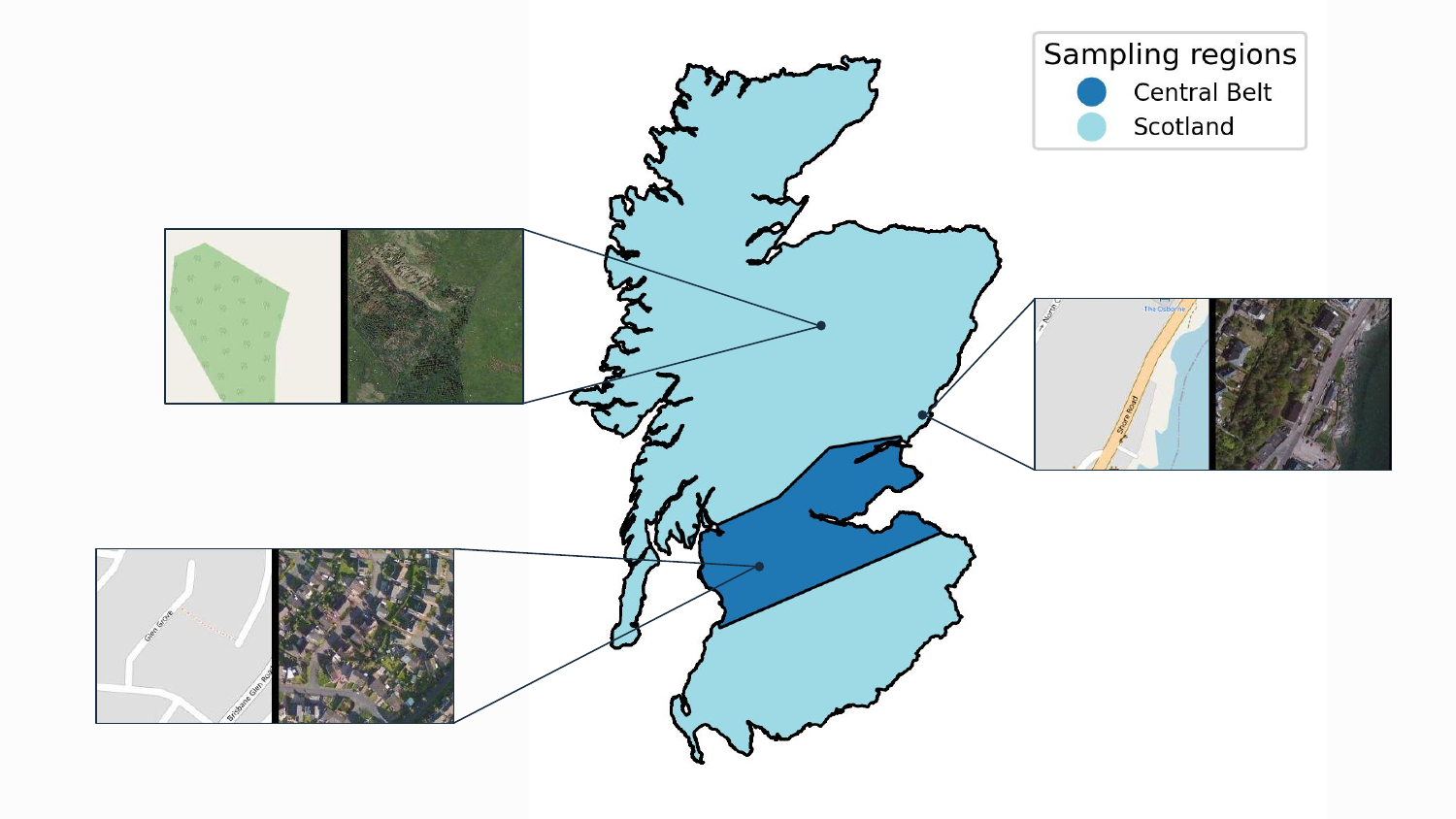}
\end{center}
   \caption{Sampling regions used for the dataset construction. We visualise some pair examples (map, satellite img). Mainland Scotland is largely rural, whereas the central belt has build up cities including Edinburgh and Glasgow.}
\label{fig:dataset}
\end{figure*}

We carry out experiments on multiple datasets, that is, sampling across different regions (Figure \ref{fig:dataset}): (1) all of Mainland Scotland, and (2) the Central Belt region. The motivation behind sampling across different regions is to account for unbalanced geographic features; Mainland Scotland is dominated by rural areas, forests, mountain ranges, and fields whereas the Central Belt region has a much larger representation of human-made structures like buildings, roads, and other features found in larger cities. The Mainland Scotland dataset contains 78,414 training pairs of images, and the Central Belt dataset 68,195 training pairs (with an additional 20\% of test pairs for each case).

We use OpenStreetMap tiles and World Imagery satellite images, both at a zoom level of 17. For the central belt region, we explore two products from the free World Imagery service as provided by ArcGis Online: the latest World Imagery version and the older Clarity version (deprecated) \cite{worldimagery_clarity}. We find that the Clarity version retains more detail and higher image quality, so we train our models on both versions for a comparative evaluation (note that World Imagery products are composites compiled from different sources and providers, resulting in varying resolutions across locations).

\section{Method}

We use the ControlNet \cite{Zhang2023Feb} architecture to train a model capable of generating realistic satellite images from OSM tiles. Before detailing the specifics of our approach, we provide a brief overview of the ControlNet method.

\subsection{ControlNet Overview}
ControlNet \cite{Zhang2023Feb} is an architecture designed to augment pretrained image diffusion models by allowing task-specific conditioning. It has the ability to manipulate the input conditions of \emph{neural network blocks}, thereby controlling the diffusion process. Intuitively, it can be seen as a way of injecting explicit guidance on the denoising process, conditioning the outputs on some reference image, in addition to the text prompt.

A~\emph{network block} in this context refers to any set of neural layers grouped as a frequently-used unit for building networks, such as a ResNet block,~\texttt{conv-bn-relu} block, and transformer block, among others.

Given a feature map ${x \in \mathbb{R}^{h\times w\times c}}$ where $\{h, w, c\}$ represent height, width, and channel numbers respectively, a neural network block $\mathcal{F}(\cdot; \theta)$ with a set of parameters $\theta$ transforms $x$ into another feature map $y$ via the relation $y = \mathcal{F}(x; \theta)$.

\begin{figure*}[h]
\begin{center}
\includegraphics[scale=0.74]{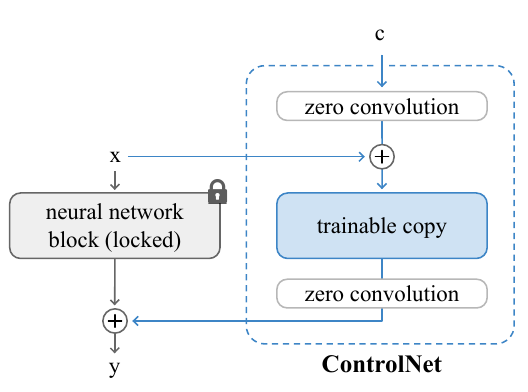}
\end{center}
   \caption{ControlNet network blocks with "zero convolutions" ($1\times1$ convolution layer with both weight and bias initialised to zeros). Figure adapted from the original work \cite{Zhang2023Feb}.}
\label{fig:controlnet}
\end{figure*}

Crucially, as Figure \ref{fig:controlnet} illustrates, ControlNet keeps the parameters $\theta$ locked, cloning it into a trainable copy $\theta_c$ which is trained with an external condition vector $c$. The idea behind making such copies instead of directly training the original weights is to mitigate overfitting risks in small datasets and being able to reuse larger models trained on billions of images.

An important innovation is the introduction of a~\emph{zero convolution} layer to connect the frozen network blocks and the trainable copies (Figure \ref{fig:controlnet}). Zero convolution is a $1\times1$ convolution layer with both weight and bias initialised as zeros. Note that ControlNet initially will not affect the original network at all, but as it is trained, it will gradually start to influence the generation with the external condition vectors.

\subsection{ControlNet for Satellite Image Synthesis}

We use the ControlNet architecture, along with a large pretrained diffusion model (Stable Diffusion) to translate OpenStreetMap images into realistic satellite images. 

We follow the same training process as in the original ControlNet architecture \cite{Zhang2023Feb}. Our model progressively denoises images in the perceptual latent space to generate samples. It learns to predict the noise added to the noisy image, and this learning objective is used in the fine-tuning process of the entire pipeline.

As the Stable Diffusion (SD) \cite{Rombach2021Dec} weights are locked, the gradient computation on the SD model can be avoided, which accelerates the training process and saves on GPU memory. Leveraging a large pretrained diffusion model not only improves computational efficiency, but also yields higher-quality results.

\subsection{Training and inference details}
We carry out multiple experiments with different pretrained large diffusion backbones. Specifically we experiment with two different versions from Stable Diffusion: v.1-5, and v.2-1. We find that SD version v.1-5 tends to give better results. Experiments are run on a cluster node of 8 A100 40GB GPUs. The batch size is set to 2048 for 250 epochs. The training time is approximately 8 hours and the learning rate is kept constant at 0.00001. During inference, images are sampled with 50 inference steps (further increasing the number of inference steps doesn't have a noticeable impact on image quality), and it takes 2-3 seconds per image.

The best performing model, trained on the Central Belt dataset, is publicly available at~\url{https://huggingface.co/mespinosami/controlearth}. We also publish the model trained on Mainland Scotland at~\url{https://huggingface.co/mespinosami/controlearth-sct}.

\section{Analysis}
\label{sec:results}

\begin{figure*}[!ht]
\vspace{1mm}
\begin{center}
\includegraphics[scale=0.5]{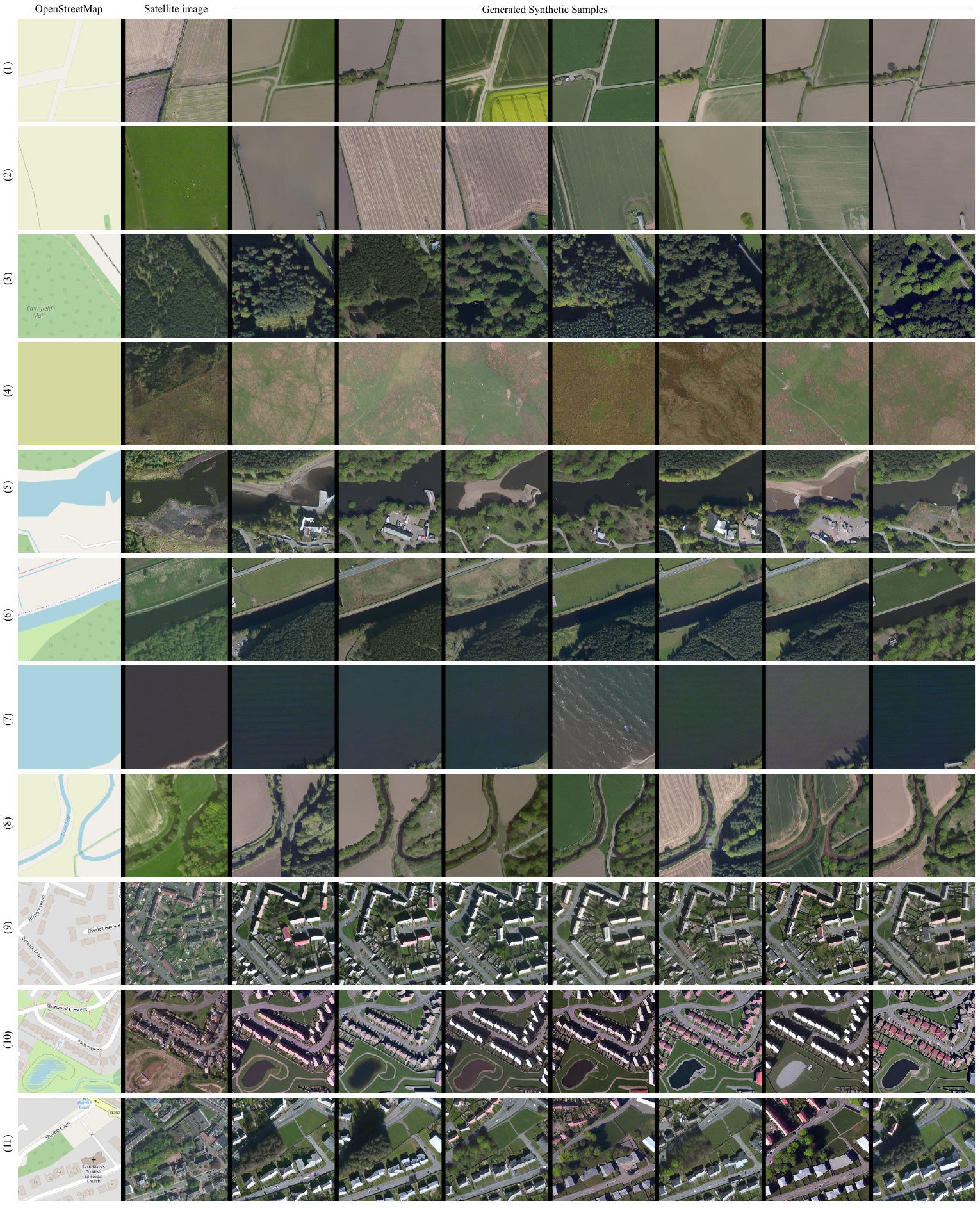}
\end{center}
   \caption{Examples of synthetic satellite images from the trained model conditioned on maps. All images shown correspond to the test set. The real satellite images are provided as reference (second column) but they are not used at inference. Rows 1-4 show agricultural land, forests and bare areas. Rows 5-8 illustrate water bodies at varying sizes. Rows 9-11 correspond to different man-made structures, which condition the generation with more intricate patterns.}
\label{fig:examples-long}
\end{figure*}

We carry out a qualitative analysis of our results, mainly involving the visual inspection of the generated satellite images. This lets us evaluate more subjective elements such as colour consistency, spatial coherence and feature representation, which are often hard to quantify.

We include a selection of successful examples in Figure \ref{fig:examples-clarity} and Figure \ref{fig:examples-long} that demonstrate the model's capabilities under different conditions (best viewed up close, in colour). 

One of the desirable behaviours that the trained model exhibits is the diversity of samples given the same map. This shows that the model has learnt to encode the variances found in the map classes (e.g. agricultural crop), thus, successfully captured the complexity of the dataset, instead of collapsing all generations to the same image. For example, rows 1-4 in Fig.~\ref{fig:examples-long} illustrate seasonality changes in the different samples. Similarly, other variances are also perceivable, such as weather phenomena, lighting conditions and human activity. Sampling with high diversity can be used as a data augmentation technique, ensuring intraclass invariance for tasks such as classification.

Rows 5-8 are examples for water bodies of multiple sizes, such as rivers, human-made canals, and open sea in coastal regions. Lastly, rows 9-11 show urban areas and more elaborate human-made patterns which the model is able to closely follow.

\begin{figure*}[!h]
\begin{center}
\includegraphics[scale=0.75]{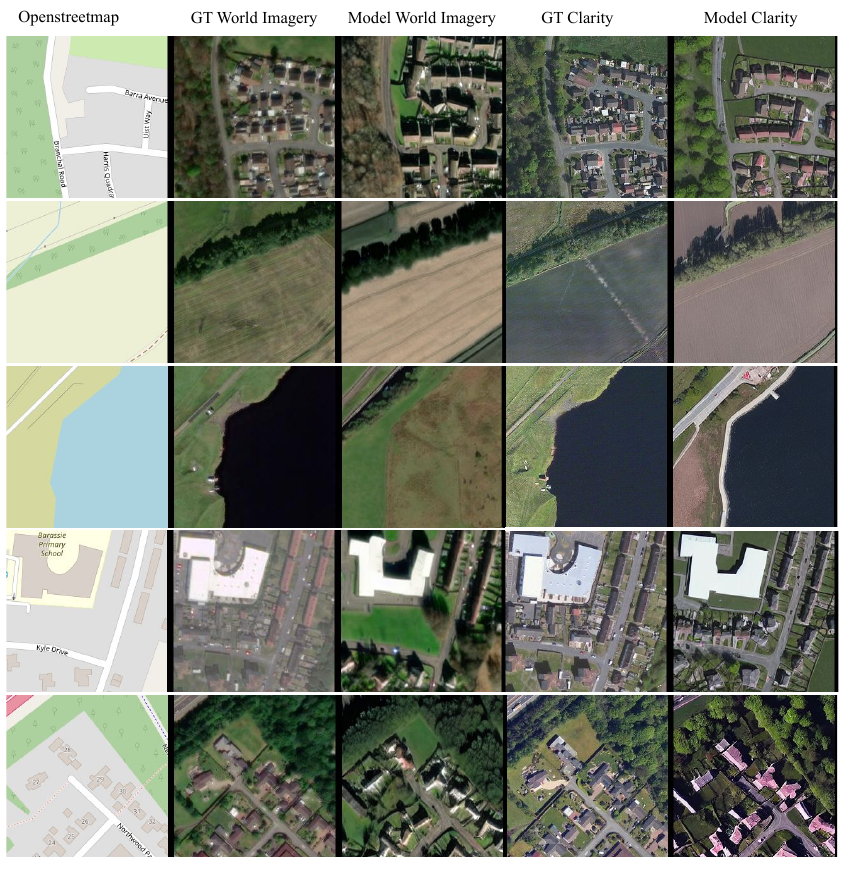}
\end{center}
   \caption{We illustrate the quality differences when training the same model with World Imagery and Clarity datasets over the Central Belt area. GT stands for Ground Truth, i.e.\ the real satellite images.}
\label{fig:comparison}
\end{figure*}

As discussed in Section \ref{sec:dataset}, we train the same model on two different versions of the central belt dataset (one with World Imagery updated product, and the other with the deprecated Clarity version). Figure \ref{fig:comparison} provides a comparative visual analysis of two identical ControlNet models, both subjected to the same training parameters but on the two distinct datasets. As it can be observed, the deprecated Clarity product shows finer details and superior image quality. Therefore, it becomes evident that the quality of the learned representations is heavily influenced by the quality of the training data employed.

\subsection{Failure cases}
\label{sec:failures}
\begin{figure*}[h]
\begin{center}
\includegraphics[scale=0.5]{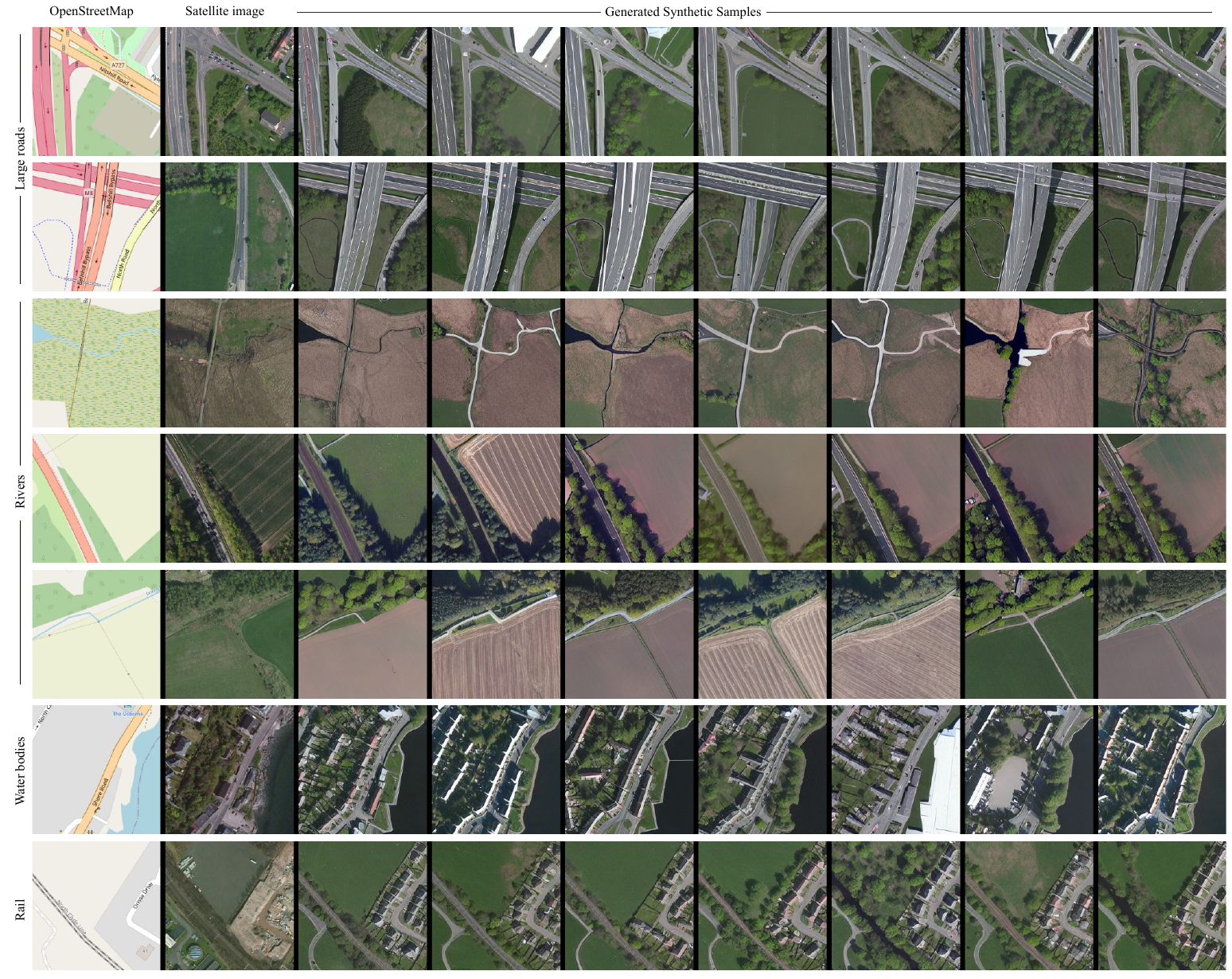}
\end{center}
   \caption{Failure cases for more challenging scenarios (which usually correspond to under-represented cases in the dataset, such as larger railways, coastal regions, or road intersections). The real satellite images are shown in the second column for comparative purposes.}
\label{fig:failure}
\end{figure*}

Some failure cases are shown in Figure \ref{fig:failure}. Large roads, specially those with lanes and straight lines are found challenging by our model. Equally, intersections and road overpasses are difficult to generate coherently. Rivers are easily mistaken by roads in some of the samples, and we show a failure case for a larger water body, where it is confused by a building (possibly due to its polygonal shape). Lastly, we also visualise railroads as challenging scenarios. These occurrences can largely be attributed to the under-representation in our dataset of the specific scenarios.

\clearpage

\section{Discussion}

The use of generative diffusion models in remote sensing still remains in its early stages. However, the results presented in this study highlight their potential.

\textit{Opportunities}: This approach allows for multiple applications. It enables the enhancement of existing datasets, by extending the number of samples. This is particularly useful for low-data regimes or scenarios where data collection can be expensive. Similarly, it can be utilised in the data augmentation step of any training pipeline. Given the diversity and realism of the generated samples it is a strong tool to ensure robustness and generalisation in models. Furthermore, the ability to synthesise high-resolution images that closely follow a specified layout (i.e.\ map) can be used to complement private datasets, providing a means to increase data accessibility without compromising confidentiality. Lastly, there exist multiple image-to-image use cases where this method could prove useful, for instance cloud or haze removal.

\textit{Challenges}: As the quality of synthesised satellite imagery improves, concerns around misuse and the propagation of fake satellite images arise. The creation of fake satellite images or its manipulation could have harmful consequences in emergency situations, or in geopolitical events. Alongside the development of this technology, there needs to be a concurrent effort on creating regulations and ethical guidelines. On the other hand, our method is capable of creating adversarial samples (i.e.\ fake satellite images that resemble realistic ones), thus, it can be leveraged to create adversarial datasets. Such datasets could be used to train models for the detection of fake or manipulated satellite imagery.

\textit{Future work:} The current method struggles with finer structures and undersampled classes (see Section \ref{sec:failures} for more details), providing room for improvement in those scenarios. Secondly, we aim to expand the current dataset by: including a wider set of modalities increasing the representational diversity (such as GIS information, DEMs, land cover data, more varied text prompts), expand its geographical coverage (to more diverse habitats and climatic regions), and develop a new sampling strategy (based on land cover maps and population density). A more complete dataset will allow for the improvement on the challenging situations across a wider range of regions. And a multi-modal dataset will enable to condition the generation process on other data modalities. Furthermore, it remains unexplored the possibilities of using different and more diverse text prompts in the generation process (for instance, for controlling seasonality changes or other weather conditions). Finally, another exciting direction is enabling consistent generation of larger maps with a smooth tiling transition. We plan to explore iterative hierarchical generation or style conditioning as possible methodologies to achieve this objective. Such method would open possibilities for artists and content creators.

\section{Conclusion}
We have demonstrated that state-of-the-art diffusion models can be used to generate realistic satellite images conditioned on maps. For this purpose, we create a large dataset containing pairs of maps and satellite images for Mainland Scotland and the Central Belt regions. With such dataset, we successfully train ControlNet models and provide insights on the results obtained. Finally, we outline some possible directions for improvements, and discuss the potential of generative methods in the field of EO.

\section{Acknowledgements}
Miguel Espinosa was supported through a Centre for Satellite Data in Environmental Science (SENSE) CDT studentship (NE/T00939X/1). This work used JASMIN, the UK's collaborative data analysis environment \url{https://jasmin.ac.uk} \cite{jasmin}. The authors are grateful to Tom Lee for his helpful comments.

\bibliography{egbib}
\end{document}